\documentclass[10pt, a4paper]{article}
\usepackage{lrec}
\usepackage{graphicx}
\usepackage{tabularx}
\usepackage{soul}
\usepackage{epstopdf}
\usepackage[utf8]{inputenc}

\usepackage{hyperref}
\usepackage{xstring}

\title{GameWikiSum: a Novel Large Multi-Document Summarization Dataset}

\name{Diego Antognini, Boi Faltings}

\address{Artificial Intelligence Laboratory \\
  École Polytechnique Fédérale de Lausanne \\
  Lausanne, Switzerland \\
  \texttt{firstname.lastname@epfl.ch}\\}

\abstract{
    Today's research progress in the field of multi-document summarization is obstructed by the small number of available datasets. 
    Since the acquisition of reference summaries is costly, existing datasets contain only hundreds of samples at most, resulting in heavy reliance on hand-crafted features or necessitating additional, manually annotated data.  
    The lack of large corpora therefore hinders the development of sophisticated models.
    Additionally, most publicly available multi-document summarization corpora are in the news domain, and no analogous dataset exists in the video game domain.
    In this paper, we propose GameWikiSum, a new domain-specific dataset for multi-document summarization, which is one hundred times larger than commonly used datasets, and in another domain than news.
    Input documents consist of long professional video game reviews as well as references of their gameplay sections in Wikipedia pages.
    We analyze the proposed dataset and show that both abstractive and extractive models can be trained on it.
    We release GameWikiSum for further research: \url{https://github.com/Diego999/GameWikiSum}. \\ \newline  }

\begin{document}

\maketitleabstract

\section{Introduction}

With the growth of the internet in the last decades, users are faced with an increasing amount of information and have to find ways to summarize it.
However, producing summaries in a multi-document setting is a challenging task; the language used to display the same information in a sentence can vary significantly, making it difficult for summarization models to capture.
Thus large corpora are needed to develop efficient models.
There exist two types of summarization: extractive and abstractive.
Extractive summarization outputs summaries in two steps, namely via sentence ranking, where an importance score is assigned to each sentence, and via the subsequent sentence selection, where the most appropriate sentence is chosen. 
In abstractive summarization, summaries are generated word by word auto-regressively, using sequence-to-sequence or language models.
Given the complexity of multi-document summarization and the lack of datasets, most researchers use extractive summarization and rely on hand-crafted features or additional annotated data, both needing human expertise. 

To our knowledge, \newcite{wiki2018} is the only work that has proposed a large dataset for multi-document summarization.
By considering Wikipedia entries as a collection of summaries on various topics given by their title (e.g., Machine Learning, Stephen King), they create a dataset of significant size, where the lead section of an article is defined as the reference summary and input documents are a mixture of pages obtained from the article's reference section and a search engine. 
While this approach benefits from the large number of Wikipedia articles, in many cases, articles contain only a few references that tend to be of the desired high quality, and most input documents end up being obtained via a search engine, which results in noisy data.
Moreover, at testing time no references are provided, as they have to be provided by human contributors.
\newcite{wiki2018} showed that in this case, generated summaries based on search engine results alone are of poor quality and cannot be used.

In contrast, we propose a novel domain-specific dataset containing $14\,652$ samples, based on professional video game reviews obtained via Metacritic\footnote{\url{https://www.metacritic.com/game}} and gameplay sections from Wikipedia.
By using Metacritic reviews in addition to Wikipedia articles, we benefit from a number of factors.
First, the set of aspects used to assess a game is limited and consequently, reviews share redundancy.
Second, because they are written by professional journalists, reviews tend to be in-depth and of high-quality.
Additionally, when a video game is released, journalists have an incentive to write a complete review and publish it online as soon as possible to draw the attention of potential customers and increase the revenue of their website \cite{Zhou2010OnlineUR}. 
Therefore, several reviews for the same product become quickly available and the first version of the corresponding Wikipedia page is usually made available shortly after.
Lastly, reviews and Wikipedia pages are available in multiple languages, which opens up the possibility for multilingual multi-document summarization.

\section{GameWikiSum}

In this section, we introduce a new domain-specific corpus for the task of multi-document summarization, based on professional video game reviews and gameplay sections of Wikipedia.

\subsection{Dataset Creation}
Journalists are paid to write complete reviews for various types of entertainment products, describing different aspects thoroughly.
Reviewed aspects in video games include the gameplay, richness, and diversity of dialogues, or the soundtrack.
Compared to usual reviews written by users, these are assumed to be of higher-quality and longer. 

Metacritic\footnote{\url{https://www.metacritic.com}} is a website aggregating music, game, TV series, and movie reviews. 
In our case, we only focus on the video game section and crawl different products with their associated links, pointing to professional reviews written by journalists.
It is noteworthy that we consider reviews for the same game released on different platforms (e.g., Playstation, Xbox) separately.
Indeed, the final product quality might differ due to hardware constraints and some websites are specialized toward a specific platform\footnote{For example \url{www.playstationlifestyle.net} has only reviews for \textit{Playstation} games.}.

Given a collection of professional reviews, manually creating a summary containing all key information is too costly at large scale as reviews tend to be long and thorough.
To this end, we analyzed Wikipedia pages for various video games and observed that most contain a gameplay section, that is an important feature in video game reviews.
Consequently, we opt for summaries describing only gameplay mechanics.
Wikipedia pages are written following the Wikipedia Manual of Style\footnote{\url{https://en.wikipedia.org/wiki/Wikipedia:Manual_of_Style}} and thus, guarantee summaries of a fairly uniform style.
Additionally, we observed that the gameplay section often cites excerpts of professional reviews, which adds emphasis to the extractive nature of GameWikiSum. 

In order to match games with their respective Wikipedia pages, we use the game title as the query in the Wikipedia search engine and employ a set of heuristic rules.

\subsection{Heuristic matching}
\label{heuristic_matching}

We crawl approximately $265\,000$ professional reviews for around $72\,000$ games and $26\,000$ Wikipedia gameplay sections.
Since there is no automatic mapping between a game to its Wikipedia page, we design some heuristics.
The heuristics are the followings and applied in this order:

\begin{enumerate}
    \item \underline{Exact title match}: titles must match exactly;
    \item \underline{Removing tags}: when a game has the same name than its franchise, its Wikipedia page has a title similar to \textit{Game (year video game)} or \textit{Game (video game)};
    \item \underline{Extension match}: sometimes, a sequel or an extension is not listed in Wikipedia. In this case, we map it to the Wikipedia page of the original game.
\end{enumerate}

We only keep games with at least one review and a matching Wikipedia page, containing a gameplay section. 
\subsection{Descriptive Statistics}

\begin{table}[h]
\small
\begin{center}
\begin{tabular}{l@{\hspace*{2mm}}c@{\hspace*{2mm}}c@{\hspace*{2mm}}c@{\hspace*{2mm}}c@{\hspace*{2mm}}c@{\hspace*{2mm}}c@{\hspace*{2mm}}}
\textbf{Percentile} & \textbf{20} & \textbf{40} & \textbf{50} & \textbf{60} & \textbf{80} & \textbf{100}\\ \hline
Num Documents & $2$ & $5$ & $7$ & $10$ & $18$ & $84$ \\
Summary Size & $139$ & $246$ & $321$ & $419$ & $684$ & $4639$ \\
Documents Size & $2\,536$ & $5\,604$ & $7\,815$ & $10\,634$ & $20\,498$ & $249\,062$ \\
ROUGE-1 recall & $67.7$ & $80.7$ & $85.29$ & $88.8$ & $94.1$ & $100.0$ \\
ROUGE-2 recall & $14.3$ & $23.0$ & $27.4$ & $31.9$ & $41.9$ & $100.0$ \\
\end{tabular}
\end{center}
\caption{\label{dataset_stats} Percentiles for different aspects of GameWikiSum. Size is in number of words. ROUGE scores are computed with a summary given its reviews.}
\end{table}

We build GameWikiSum corpus by considering English reviews and Wikipedia pages.
Table~\ref{dataset_stats} describes its overall properties.
Most samples contain several reviews, whose cumulative size is too large for extractive or abstractive models to be trained in an end-to-end manner. The total vocabulary is composed of $282\,992$ words. Our dataset also comes from a diverse set of sources: over $480$ video game websites appear as source documents in at least $6$ video games; they are responsible for $99.95\%$ of the reviews.

Following \newcite{wiki2018}, a subset of the input has to be therefore first coarsely selected, using extractive summarization, before training an extractive or abstractive model that generates the Wikipedia gameplay text while conditioning on this extraction. Additionally, half of the summaries contain more than three hundred words (see Table~\ref{dataset_stats}), which is larger than previous work.

To validate our hypothesis that professional game reviews focus heavily on gameplay mechanics, we compute the proportion of unigrams and bigrams of the output given the input.
We observe a significant overlap ($20\%$ documents containing $67.7\%$ of the words mentioned in the summary, and at least $27.4\%$ bigrams in half of the documents), emphasizing the extractive nature of GameWikiSum.
Several examples of summaries are shown in Section~\ref{sec_examples}

\begin{table*}
\begin{center}
\begin{tabular}{lcccc}
\textbf{Dataset} & \textbf{Input} & \textbf{Output} & \textbf{\# Examples} & \textbf{ROUGE-1 R}\\ \hline
*Gigaword \cite{graff2003} & $10^1$ & $10^1$ & $10^6$ & $78.7$\\
**CNN/DailyMail \cite{nallapati2016abstractive} & $10^2$-$10^3$ & $10^1$ & $10^5$ & $76.1$\\
DUC 2001-2004\footnotemark & $10^3$ & $10^2$ & $10^2$ & $94.4$ \\
TAC 2008-2011\footnotemark & $10^3$ & $10^2$ & $10^2$ & $95.3$ \\
WikiSum \cite{wiki2018} & $10^2$-$10^6$ & $10^1$-$10^3$ & $10^6$ & $59.2$\\
GameWikiSum (ours) & $10^3$-$10^5$ & $10^2$-$10^3$ & $10^4$ & $80.1$
\end{tabular}
\end{center}
\caption{\label{dataset_comp} Sizes and unigram recall of single (marked with *) and multi-document summarization datasets. Recall is computed with reference summaries given the input documents.}
\end{table*}
\addtocounter{footnote}{-1}
\footnotetext{\url{https://www-nlpir.nist.gov/projects/duc/guidelines.html}}
\addtocounter{footnote}{+1}
\footnotetext{\url{http://www.nist.gov/tac/}}

Table~\ref{dataset_comp} shows a comparison between GameWikiSum and other single and multi-document summarization datasets.
GameWikiSum has larger input and output size than single document summarization corpora (used in extractive and abstractive models) while sharing similar word overlap ratios.
Compared to DUC and TAC (news domain), GameWikiSum is also domain-specific and has two orders of magnitude more examples, facilitating the use of more powerful models. 
Finally, WikiSum has more samples but is more suitable for general abstractive summarization, as its articles cover a wide range of areas and have a lower word overlap ratio.

We divide GameWikiSum into train, validation and testing sets with a rough ratio of 80/10/10, resulting in $11\,744$, $1\,454$ and $1\,454$ examples respectively.
If a game has been released on several platforms (represented by different samples), we group them in the same subset to avoid review overlap between training, validation, and testing. The distribution of samples per platform is shown in Table~\ref{game_distr}. We compute in addition the mean number of input documents, ROUGE-1, and ROUGE-2 scores of the output given the input. We observe that most platforms have a mean ROUGE-1 score above $80$ and $30$ for ROUGE-2.

\begin{table*}
\begin{center}
  \begin{tabular}{lrccc}
    \textbf{Platform} & \textbf{\# Games} & \textbf{\# Documents} & \textbf{ROUGE-1 R} & \textbf{ROUGE-2 R}\\
        \hline
PC & $3586$ & $8 \pm 8$ & $81.18 \pm 15.45$ & $27.32 \pm 14.52$\\
Wii U & $224$ & $10 \pm 13$ & $86.47 \pm 10.78$ & $34.14 \pm 16.03$\\
Nintendo 64 & $66$ & $8 \pm 3$ & $77.46 \pm 13.10$ & $21.11 \pm 9.37$\\
Dreamcast & $83$ & $6 \pm 2$ & $66.12 \pm 13.73$ & $13.01 \pm \ 6.27$\\
PlayStation & $86$ & $4 \pm 2$ & $60.95 \pm 14.67$ & $10.97 \pm \ 6.47$\\
PlayStation 2 & $954$ & $13 \pm 9$ & $85.93 \pm 11.74$ & $30.47 \pm 11.89$\\
Game Boy Advance & $368$ & $5 \pm 4$ & $69.38 \pm 17.78$ & $17.23 \pm 11.15$\\
GameCube & $341$ & $10 \pm 7$ & $82.26 \pm 12.16$ & $24.95 \pm 10.66$\\
Xbox & $486$ & $15 \pm 9$ & $88.40 \pm \ 9.95$ & $32.31 \pm 10.79$\\
DS & $679$ & $10 \pm 9$ & $85.27 \pm 11.77$ & $30.99 \pm 13.38$\\
PSP & $407$ & $12 \pm 9$ & $85.08 \pm 13.85$ & $30.71 \pm 13.27$\\
Xbox 360 & $1358$ & $19 \pm 14$ & $86.90 \pm 14.54$ & $34.93 \pm 15.72$\\
PlayStation 3 & $1128$ & $13 \pm 11$ & $84.53 \pm 16.27$ & $32.28 \pm 15.48$\\
Wii & $665$ & $10 \pm 10$ & $84.70 \pm 14.07$ & $32.18 \pm 14.77$\\
iOS & $1344$ & $4 \pm 3$ & $77.86 \pm 15.48$ & $23.39 \pm 13.26$\\
Xbox One & $817$ & $8 \pm 9$ & $83.33 \pm 14.53$ & $30.66 \pm 15.63$\\
3DS & $312$ & $15 \pm 14$ & $88.62 \pm 12.87$ & $39.75 \pm 19.01$\\
PlayStation Vita & $337$ & $7 \pm 9$ & $80.97 \pm 14.50$ & $28.21 \pm 16.63$\\
PlayStation 4 & $1103$ & $14 \pm 14$ & $87.42 \pm 14.02$ & $37.84 \pm 18.00$\\
Switch & $308$ & $11 \pm 12$ & $89.97 \pm \ 9.64$ & $38.61 \pm 15.95$\\
\hline
All & $14652$ & $11 \pm 11$ & $83.19 \pm 15.04$ & $29.99 \pm 15.48$\\

    \end{tabular}
    \end{center}
  \caption{\label{game_distr}Game distribution over platforms with their average and standard deviation number of input documents and ROUGE scores.}
\end{table*}

\section{Experiments and Results}

\subsection{Evaluation Metric}
\label{sub:evaluation_metric}

We use the standard ROUGE \cite{Lin2004} used in summarization and report the ROUGE-L F1 score.
ROUGE-L F1 is more appropriate to measure the quality of generated summaries in this context because summary lengths are longer than usual (see Table~\ref{dataset_comp}) and vary across the dataset (see Table~\ref{dataset_stats}).
Another motivation to use ROUGE-L F1 is to compare abstractive models with extractive ones, as the output length is unknown a priori for the former, but not for the latter. 
We report in addition ROUGE-1 and ROUGE-2 recall scores.

To ensure consistent results across all comparative experiments, extractive models generate summaries of the same length as reference summaries.
In realistic scenarios, summary lengths are not pre-defined and can be adjusted to produce different types of summaries (e.g., short, medium or long).
We do not explicitly constrain the output length for abstractive models, as each summary is auto-regressively generated.

\subsection{Baselines}

For extractive models, we include LEAD-$k$ which is a strong baseline for single document summarization tasks and takes the first $k$~sentences in the document as summary \cite{see2017get}.
TextRank~\cite{Mihalcea04TextRank} and LexRank~\cite{erkan2004lexrank} are two graph-based methods, where nodes are text units and edges are defined by a similarity measure.
SumBasic~\cite{nenkova2005impact} is a frequency-based sentence selection method, which uses a component to re-weigh the word probabilities in order to minimize redundancy.
The last extractive baselines are the near state-of-the-art models C\_SKIP from~\newcite{rossiello2017centroid} and SemSenSum from~\newcite{antognini2019}.
The former exploits the capability of word embeddings to leverage semantics, whereas the latter aggregates two types of sentence embeddings using a sentence semantic relation graph, followed by a graph convolution.

We use common abstractive sequence-to-sequence baselines such as Conv2Conv~\cite{gehring2017convolutional}, Transformer~\cite{vaswani2017attention} and its language model variant, TransformerLM~\cite{wiki2018}.
We use implementations from \textit{fairseq}\footnote{\href{https://github.com/pytorch/fairseq}{github.com/pytorch/fairseq}} and \textit{tensor2tensor}\footnote{\href{https://github.com/tensorflow/tensor2tensor}{github.com/tensorflow/tensor2tensor}}.
As the corpus size is too large to train extractive and abstractive models in an end-to-end manner due to hardware constraints, we use Tf-Idf to coarsely select sentences before training similarly to \newcite{wiki2018}.
We limit the input size to $2$K tokens so that all models can be trained on a Titan Xp GPU (12GB GPU RAM).
We run all models with their best reported parameters.

\subsection{Results}

\begin{table}[h]
  \centering
  \begin{tabular}{lccc}
    \textbf{Model} & R-L & R-1 & R-2 \ \\
        \hline
 LEAD-3 & $11.45$& $12.77$ & $2.45$ \\
 LEAD-5 & $18.78$ & $19.82$ & $3.42$ \\
 TextRank & $29.30$ & $31.07$ & $4.96$\\
 LexRank & $29.74$ & $31.26$ & $4.96$\\
 SumBasic & $30.36$ & $31.82$ & $4.79$\\
 C\_SKIP & $31.66$ & $32.90$ & $5.25$\\
 SemSenSum & $31.72$ & $35.11$ & $5.56$\\
 Conv2Conv* & $20.10$ & $19.30$ & $5.20$\\
 Transformer* & $14.60$ & $16.00$ & $2.80$\\
 TransformerLM* &  $\ 9.52$ & $\ \ 7.03$ & $1.17$\\
    \end{tabular}
  \caption{\label{sum_perf}Comparison extractive and abstractive (marked with *) models. Reported scores correspond to ROUGE-L F1 score, ROUGE-1 and ROUGE-2 recall respectively.}
\end{table}

Table~\ref{sum_perf} contains the results.
LEAD-$5$ achieves less than $20$ for ROUGE-L as well as ROUGE-1 and less than $3.5$ for ROUGE-2. Taking only $3$ sentences leads to even worse results: below $13$ and $3$ respectively.
Unlike in other datasets, these results are significantly outperformed by all other extractive models but surprisingly, abstractive models perform worse on average. This demonstrates the difficulty of the task in GameWikiSum compared to \newcite{nallapati2016abstractive} and \newcite{graff2003}.

For extractive models, TextRank and LexRank perform worse than other models. 
The frequency-based model SumBasic performs slightly better but does not achieve comparable results with embedding-based models.
Best results are obtained with C\_SKIP and SemSentSum, showing that more sophisticated models can be trained on GameWikiSum and improve results significantly.
Interestingly, taking into account the context of a sentence and hence better capturing the semantics, SemSentSum achieves only slightly better scores than C\_SKIP, which relies solely on word embedding.
We show in Section~\ref{sec_examples} several examples with their original summaries and generated ones with the best model.

Overall, the abstractive performance of sequence-to-sequence and language models are significantly lower than C\_SKIP and SemSentSum in terms of ROUGE-L and ROUGE-1.
However, Conv2Conv obtains only $0.05$ less ROUGE-2 score compared to C\_SKIP and $0.36$ to SemSentSum. We suspect ROUGE-2 to be easier for abstractive sequence-to-sequence models, as half of the samples only have a ROUGE-2 around $27.00$ without any limitation of the input size (see Table~\ref{dataset_stats}).
Consequently, copying sentences from a small subset of the whole input documents for extractive models leads to worse ROUGE-2 recall. 
A normal transformer underperforms compared to Conv2Conv, and its language model variant achieves significantly worse results than other models due to a lack of data.

We highlight that GameWikiSum has two orders of magnitude fewer samples (see Table~\ref{dataset_comp}) compared to \newcite{wiki2018}. Therefore, it is necessary to have either additional annotated data or pre-train TransformerLM on another corpus.

\subsection{Examples of Original and Generated Summaries}
\label{sec_examples}

\begin{figure*}
\center
  \includegraphics[width=.99\linewidth]{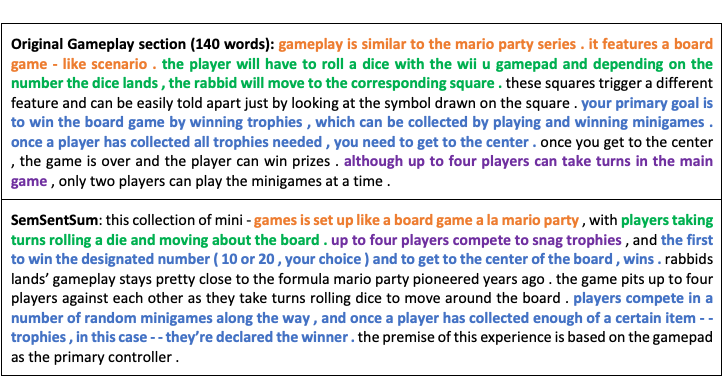}\\
  \includegraphics[width=.99\linewidth]{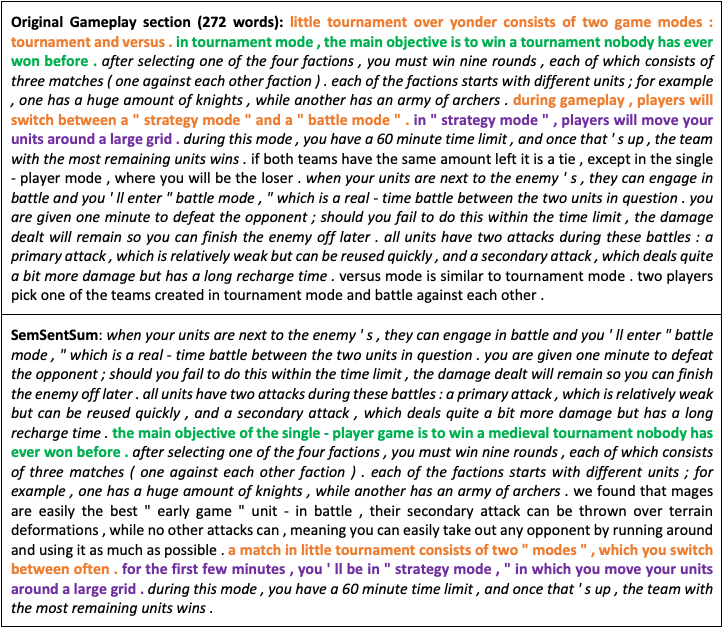}
  \caption{\label{fig:examples}Reference summary and a summary produced by SemSentSum for the game \textit{Rabbids Land}\protect\footnotemark{} and \textit{Little Tournament Over Yonder}\protect\footnotemark{}. Sentences in italic correspond to identical text fragments and colored ones describe the same content.}
\end{figure*}

Figure~\ref{fig:examples} shows two samples with their gameplay sections from Wikipedia and summaries generated by the best baseline SemSentSum.
In the first example, we notice that the model has selected sentences from the reviews that are also in the original Wikipedia page.
Additionally, we observe, for both examples, that several text fragments describe the same content with different sentences.
Consequently, this supports our hypothesis that professional reviews can be used in a multi-document summarization setting to produce summaries reflecting the gameplay section of Wikipedia pages.

\addtocounter{footnote}{-1}
\footnotetext{\url{https://en.wikipedia.org/wiki/Rabbids_Land}}
\addtocounter{footnote}{+1}
\footnotetext{\url{https://en.wikipedia.org/wiki/Little_Tournament_Over_Yonder}}

\section{Related Work}

To the best of our knowledge, DUC and TAC are the first multi-document summarization datasets.
They contain documents about the same event and human-written summaries.
Unsurprisingly, this approach does not scale and they could only collect hundreds of samples as shown in Table~\ref{dataset_comp}.

\newcite{zopf2016next} applied a similar strategy using Wikipedia, where they asked annotators to first tag and extract information nuggets from the lead section of Wikipedia articles.
In a further step, the same annotators searched for source documents using web search engines.
As the whole process depends on humans, they could only collect around one thousand samples.
Other attempts such as \cite{cao2016tgsum} have been made using Twitter, but the resulting dataset size was even smaller.

Only the recent work of \newcite{wiki2018} addresses the automatic creation of a large-scale multi-document summarization corpus, WikiSum.
Summaries are lead sections of Wikipedia pages and input documents a mixture of 1) its citations from the reference section 2) results from search engines using the title of the Wikipedia page as the query.
However, references (provided by contributors) are needed for their model to generate lead sections which are not garbaged texts, as shown in the experiments \cite{wiki2018}.
Consequently, this dataset is unusable for real use-cases. Similarly, \newcite{zopf-2018-auto} propose a multilingual Multi-Document dataset of approximately $7\,000$ examples based on English and German Wikipedia articles. We, however, are focused on the video game domain and provide twice more samples.

\section{Conclusion}

In this work, we introduce a new multi-document summarization dataset, GameWikiSum, based on professional video game reviews, which is one hundred times larger than commonly used datasets.
We conclude that the size of GameWikiSum and its domain-specificity makes the training of abstractive and extractive models possible.
In future work, we could increase the dataset with other languages and use it for multilingual multi-document summarization. We release GameWikiSum for further research: \url{https://github.com/Diego999/GameWikiSum}.

\section{Bibliographical References}
\label{main:ref}

\bibliographystyle{lrec}

\end{document}